\documentclass[final,onecolumn,1p]{elsarticle} 

\usepackage{amsmath}
\usepackage{amsthm}
\usepackage{algorithm}
\usepackage{algorithmic}
\usepackage{amsfonts}
\usepackage{multirow} 
\usepackage{diagbox} 
\usepackage[figuresright]{rotating} 
\usepackage{threeparttable} 
\usepackage[colorlinks=true,linkcolor=red,urlcolor=blue, citecolor= green, anchorcolor=blue]{hyperref} 
\usepackage{color}
\usepackage{cuted}
\usepackage{capt-of} 
\usepackage{dsfont}
\usepackage{listings}
\usepackage{xcolor}
\usepackage{float}
\definecolor{codegreen}{rgb}{0,0.6,0}
\definecolor{codegray}{rgb}{0.5,0.5,0.5}
\definecolor{codepurple}{rgb}{0.58,0,0.82}
\definecolor{backcolour}{rgb}{0.95,0.95,0.92}
\lstdefinestyle{mystyle}{
    backgroundcolor=\color{backcolour},   
    commentstyle=\color{codegreen},
    keywordstyle=\color{magenta},
    numberstyle=\tiny\color{codegray},
    stringstyle=\color{codepurple},
    basicstyle=\ttfamily\footnotesize,
    breakatwhitespace=false,         
    breaklines=true,                 
    captionpos=b,                    
    keepspaces=true,                 
    numbers=left,                    
    numbersep=5pt,                  
    showspaces=false,                
    showstringspaces=false,
    showtabs=false,                  
    tabsize=2}
\usepackage{graphicx}
\usepackage{subcaption}
\usepackage{amsmath}

\begin{document}

\begin{frontmatter}
\title{Understanding Diffusion Models via Code Execution}

\author[a,b]{Cheng Yu}
\ead{disanda@cqut.edu.cn}

\address[a]{School of Artificial Intelligence, Chongqing University of Technology, Chongqing , China}

\address[b]{DiAi Corporation, China}

\begin{abstract}
Diffusion models have achieved remarkable performance in generative modeling, yet their theoretical foundations are often intricate, and the gap between mathematical formulations in papers and practical open-source implementations can be difficult to bridge. Existing tutorials primarily focus on deriving equations, offering limited guidance on how diffusion models actually operate in code. To address this, we present a concise implementation of approximately 300 lines that explains diffusion models from a code-execution perspective. Our minimal example preserves the essential components---including forward diffusion, reverse sampling, the noise-prediction network, and the training loop---while removing unnecessary engineering details. This technical report aims to provide researchers with a clear, implementation-first understanding of how diffusion models work in practice and how code and theory correspond. Our code and pre-trained models are available at: \url{https://github.com/disanda/GM/tree/main/DDPM-DDIM-ClassifierFree}.

\end{abstract}


\end{frontmatter}

\section{Introduction}

We categorize modern generative models into five major families (see Fig. \ref{fig1}): generative adversarial networks (GANs) \cite{GAN}, attention-based autoregressive transformers (e.g., GPTs) \cite{Attention}, variational autoencoders (VAEs) \cite{VAE}, normalizing flows \cite{glow}, and diffusion models \cite{Score-Based,ddpm,ddim,classifier-free-diffusion-guidance}. Among these categories, GANs and diffusion models have demonstrated particularly strong performance on image and video generation tasks \cite{stable-diffusion,LLF,RFM,MTA}. The diffusion family can be traced back to early score-based generative modeling \cite{Score-Based}, which later evolved into the widely recognized Denoising Diffusion Probabilistic Models (DDPM) \cite{ddpm}. Following the extensive line of GAN-based research, diffusion models have rapidly become the dominant paradigm in computer vision due to their superior image synthesis quality and their strong ability to align with text and other modalities.

\begin{figure}[hbt]
  \centering
  \includegraphics[width=0.9\linewidth]{figs/fig1} 
  \caption{A conceptual illustration of five major generative modeling paradigms, shown from top to bottom: autoregressive transformers (GPT-style models), normalizing flows, generative adversarial networks (GANs), variational autoencoders (VAEs), and diffusion models. The figure highlights the core principles of each family and their distinctive generation mechanisms.}
  \label{fig1}
\end{figure}

However, compared with other generative paradigms, diffusion models impose a substantially higher theoretical and implementation barrier. Building a diffusion-based system such as Stable Diffusion \cite{stable-diffusion} is considerably more challenging than implementing GAN-based models like StyleGAN \cite{StyleGAN}. To bridge this gap, this work explains diffusion models from an execution-centric perspective. We provide a concise, approximately 300-line Python implementation that reproduces three representative diffusion models based on the DDPM framework, aiming to make diffusion modeling more accessible to researchers and practitioners.

\section{Related Work}

\textbf{DDPM}. Denoising Diffusion Probabilistic Models (DDPM) \cite{ddpm} formulate generative modeling as a Markovian denoising process that gradually removes noise from a normally distributed latent variable. DDPM provides a principled likelihood-based framework and achieves high-quality synthesis when trained with a sufficiently expressive neural network. In this work, we adopt the standard DDPM training objective, perform full-resolution denoising inference, and use the resulting pretrained DDPM models as the backbone for subsequent sampling and conditioning techniques.

\textbf{DDIM}. Denoising Diffusion Implicit Models (DDIM) \cite{ddim} introduce a deterministic and non-Markovian sampling procedure that significantly accelerates inference while preserving sample quality. Unlike DDPM’s stochastic reverse process, DDIM defines an implicit generative trajectory that enables fast sampling with substantially fewer steps. Building on our pretrained DDPM models, we utilize DDIM as a fast inference alternative, allowing efficient sample generation without retraining the model.

\textbf{CFG}. Classifier-free diffusion guidance (CFG) \cite{classifier-free-diffusion-guidance} enables controllable generation by jointly training conditional and unconditional diffusion models. During inference, the guidance signal is obtained by combining both predictions, providing strong alignment with user-specified conditions (e.g., text prompts or class labels). In our framework, we incorporate CFG on top of DDPM- and DDIM-based samplers, enabling conditional generation while maintaining consistent image quality and efficient inference.

\section{DDPM}

The DDPM experiments were successfully tested on both 2080 Ti and 3090 GPUs, while inference can be performed without a GPU. We recommend using Python version 3.7 or above, and PyTorch version 1.6 or higher. The training datasets include MNIST, Fashion-MNIST, and CIFAR-10.

\subsection{Parameter Setup for Noising}

\textbf{Timesteps.} 
The number of forward noising steps is denoted as \texttt{timesteps}, with a default value of 1000. Each integer value $t \in \{0, \dots, \texttt{timesteps}\}$ represents a diffusion timestep.

A larger number of steps generally results in a smoother and more stable diffusion trajectory. In our MNIST experiments, using more than 100 steps is already sufficient to ensure reasonable performance. Importantly, during training, the timestep $t$ is sampled \textbf{randomly} and does \textbf{not} follow any temporal order; it is drawn as a random integer within $[0, \texttt{timesteps}]$. In this work, we set $\texttt{timesteps} = 300$.

\vspace{5pt}
\textbf{$\mathbf{\alpha}$ schedule.} 
The sequence $\{\alpha_t\}_{t=1}^{T}$ consists of values slightly below 1 and monotonically decreasing. There are \texttt{timesteps} elements in total, and each $\alpha_t$ denotes the value at timestep $t$. In this work, $\alpha_t$ ranges from $[0.9997, 0.97]$.

\vspace{5pt}
\textbf{$\mathbf{\beta}$ schedule.} 
The sequence $\{\beta_t\}_{t=1}^{T}$ is defined as 
\[
\beta_t = 1 - \alpha_t,
\] 
forming a monotonically increasing sequence with values close to zero. Each $\beta_t$ corresponds to timestep $t$. In our implementation, $\beta_t \in [0.0003, 0.03]$.

\vspace{5pt}
\textbf{Cumulative product $\bar{\alpha}$.} 
The sequence $\{\bar{\alpha}_t\}$ is the cumulative product of $\alpha$:
\[
\bar{\alpha}_t = \prod_{s=1}^{t} \alpha_s.
\]
It forms a decreasing sequence from values near 1 to values approaching 0 (excluding both boundaries).

~\\
\paragraph{Example:} For $\alpha = [0.9, 0.8, 0.7]$ (i.e., $\texttt{timesteps} = 3$):
\[
\bar{\alpha} = [0.9,\, 0.9 \cdot 0.8,\, 0.9 \cdot 0.8 \cdot 0.7] = [0.9,\, 0.72,\, 0.504].
\]
The value $\bar{\alpha}_t$ simply denotes the entry at timestep $t$.

\vspace{5pt}
\textbf{$\sqrt{1 - \bar{\alpha}}$ schedule.} 
Complementary to $\bar{\alpha}$, the sequence $\sqrt{1 - \bar{\alpha}_t}$ is monotonically increasing from 0 toward 1 (excluding both boundaries). This term is directly used in the closed-form forward diffusion equation.

This differs from the cumulative $\bar{\beta}$ (the cumulative product of $\beta$), which forms a sequence decreasing toward 0 and is not used in our implementation.

\vspace{5pt}
\textbf{Image samples $x_0$.} 
The clean data samples from the training dataset serve as $x_0$, i.e., the original uncorrupted images.

\vspace{5pt}
\textbf{Gaussian noise $\epsilon$.} 
The noise added in the forward process is drawn from a standard normal distribution:
\[
\epsilon \sim \mathcal{N}(0, 1),
\]
where $\epsilon$ is a high-dimensional tensor with the same shape as $x_0$.

\subsection{Forward Diffusion Equation}

The forward noising process is modeled as a Markov chain:

\[
x_t = \sqrt{\alpha_t} \, x_{t-1} + \sqrt{1 - \alpha_t} \, \epsilon,
\]

where $\epsilon \sim \mathcal{N}(0,1)$ is Gaussian noise. This equation describes a stepwise Gaussian perturbation from $x_0$ to $x_t$, applied for $t = 1, \dots, \texttt{timesteps}$. The cumulative term $\bar{\alpha}_t$ controls the relative contribution of the original image and the noise:

\[
x_t \sim \mathcal{N}(\sqrt{\bar{\alpha}_t} \, x_0, (1 - \bar{\alpha}_t) \, \mathbf{I}),
\]

where $\bar{\alpha}_t = \prod_{s=1}^{t} \alpha_s$ is the cumulative product of $\alpha$ up to timestep $t$. 

As $t$ increases, $\alpha_t$ decreases, and the proportion of the original signal in $x_t$ diminishes while the noise proportion grows. Eventually, $x_t$ approaches pure Gaussian noise.

Directly iterating through all timesteps during training is inefficient (similar to an RNN). Fortunately, a simplified form allows us to sample $x_t$ directly from $x_0$ without simulating all intermediate steps:

\[
x_t = \sqrt{\bar{\alpha}_t} \, x_0 + \sqrt{1 - \bar{\alpha}_t} \, \epsilon.
\]

In practice, for each training batch, we randomly sample a timestep $t$ for each image, compute $x_t$ using the above equation, and use it to train the model by minimizing the mean squared error (MSE) between the predicted and true noise.

\subsection{Forward Diffusion Process}

\subsubsection{Data Preprocessing}

During training, each batch of images (batch size = $B$) is assigned a random timestep $t$ drawn uniformly from $[0, \texttt{timesteps}]$. This can be implemented in PyTorch as:

\begin{lstlisting}[language=Python]
t = torch.randint(0, timesteps, (batch_size,), device=device).long()
\end{lstlisting}

Next, we compute the noise schedule parameters $\beta$, $\alpha$, and their cumulative product $\bar{\alpha} = \prod_{s=1}^{t} \alpha_s$:

\begin{lstlisting}[language=Python]
def linear_beta_schedule(timesteps):
    scale = 1000 / timesteps
    beta_start = 0.0003 * scale
    beta_end = 0.03 * scale
    return torch.linspace(beta_start, beta_end, timesteps, dtype=torch.float64)

betas = linear_beta_schedule(timesteps)
alphas = 1.0 - betas
alphas_cumprod = torch.cumprod(alphas, axis=0)
\end{lstlisting}

The corresponding square root coefficients are computed as:

\begin{lstlisting}[language=Python]
sqrt_alphas_cumprod = torch.sqrt(alphas_cumprod)
sqrt_one_minus_alphas_cumprod = torch.sqrt(1.0 - alphas_cumprod)
\end{lstlisting}

To index the coefficients for a batch of timesteps, we define:

\begin{lstlisting}[language=Python]
def _extract(a: torch.FloatTensor, t: torch.LongTensor, x_shape):
    batch_size = t.shape[0]
    out = a.to(t.device).gather(0, t).float()
    out = out.reshape(batch_size, *((1,) * (len(x_shape) - 1)))
    return out
\end{lstlisting}

The forward diffusion sample $x_t$ is then obtained using:

\begin{lstlisting}[language=Python]
def q_sample(x_start: torch.FloatTensor, t: torch.LongTensor, noise=None):
    sqrt_alphas_cumprod_t = _extract(sqrt_alphas_cumprod, t, x_start.shape)
    sqrt_one_minus_alphas_cumprod_t = _extract(sqrt_one_minus_alphas_cumprod, t, x_start.shape)
    return sqrt_alphas_cumprod_t * x_start + sqrt_one_minus_alphas_cumprod_t * noise
\end{lstlisting}

\paragraph{Example}: For a batch of size $B=5$, a random sample of timesteps might be:

\[
t = [75, 112, 268, 207, 90]
\]

The corresponding $\sqrt{\bar{\alpha}_t}$ values after indexing and reshaping are:

\begin{verbatim}
sqrt_alphas_cumprod_t =
tensor([[[[0.5979]]],
       [[[0.3268]]],
       [[[0.0018]]],
       [[[0.0234]]],
       [[[0.4814]]]])
\end{verbatim}

These are broadcasted over the image dimensions so that each pixel of the sample receives the correct noise scaling.

\subsubsection{Training Objective}

To train the model, we sample Gaussian noise $\epsilon \sim \mathcal{N}(0,1)$ and compute $x_t$ for each image:

\[
x_t = \sqrt{\bar{\alpha}_t} \, x_0 + \sqrt{1 - \bar{\alpha}_t} \, \epsilon
\]

The U-Net model predicts $\epsilon_\theta$ given $x_t$ and $t$. The loss is the mean squared error (MSE) between the predicted and true noise:

\begin{lstlisting}[language=Python]
def train_losses(model, x_start: torch.FloatTensor, t: torch.LongTensor):
    noise = torch.randn_like(x_start)
    x_noisy = q_sample(x_start, t, noise=noise)
    predicted_noise = model(x_noisy, t)
    loss = F.mse_loss(noise, predicted_noise)
    return loss
\end{lstlisting}

This completes the forward diffusion process and defines the training objective for DDPM.

\section{DDPM Denoising}

The denoising stage is the generative phase (inference) of a DDPM, where pure Gaussian noise 
is gradually transformed into a clean image. In contrast to the training phase, 
inference must proceed in reverse order, from $x_T$ to $x_0$, and DDPM sampling 
cannot skip timesteps.

\subsection{Parameter Definitions}

\textbf{Cumulative Product $\bar{\alpha}_{t-1}$}. The quantity $\bar{\alpha}_{t-1}$ is obtained by padding the cumulative 
product of $\alpha$ with a leading 1:

\[
\bar{\alpha}_{t-1} = 
\text{pad}\big(\bar{\alpha}_{0:t-1},\ 1\big).
\]

It is used to compute the posterior distribution $q(x_{t-1} \mid x_t, x_0)$, code as:

\begin{lstlisting}[language=Python]
alphas_cumprod_prev = F.pad(self.alphas_cumprod[:-1], (1, 0), value=1.)
\end{lstlisting}

\vspace{5pt}
\textbf{Posterior Mean $\mu_{t-1}$}. The posterior mean is given by

\[
\mu_{t-1} =
\frac{\sqrt{\bar{\alpha}_{t-1}}\beta_t}{1 - \bar{\alpha}_t} x_0
\;+\;
\frac{(1 - \bar{\alpha}_{t-1})\sqrt{\alpha_t}}{1 - \bar{\alpha}_t} x_t.
\]

We denote the coefficients:

\[
\text{posterior\_mean\_coef1} =
\frac{\sqrt{\bar{\alpha}_{t-1}}\beta_t}{1-\bar{\alpha}_t},
\qquad
\text{posterior\_mean\_coef2} =
\frac{(1-\bar{\alpha}_{t-1})\sqrt{\alpha_t}}{1-\bar{\alpha}_t}.
\]

\begin{lstlisting}[language=Python]
posterior_mean_coef1 = betas * torch.sqrt(self.alphas_cumprod_prev) / (1.0 - alphas_cumprod)
posterior_mean_coef2 = (1.0 - alphas_cumprod_prev) * torch.sqrt(alphas) / (1.0 - alphas_cumprod)
\end{lstlisting}

Posterior Variance:

\[
\sigma_{t-1}^2 = 
\frac{\beta_t(1-\bar{\alpha}_{t-1})}{1 - \bar{\alpha}_t}.
\]

\begin{lstlisting}[language=Python]
posterior_variance = betas * (1.0 - alphas_cumprod_prev) / (1.0 - alphas_cumprod)
\end{lstlisting}

To avoid numerical underflow, the logarithm is stored:

\begin{lstlisting}[language=Python]
posterior_log_variance_clipped = torch.log(self.posterior_variance.clamp(min=1e-20))
\end{lstlisting}

\textbf{Estimating $x_0$}. The reverse of the forward noising process:

\[
x_0 = 
\frac{x_t}{\sqrt{\bar{\alpha}_t}}
-
\frac{\sqrt{1 - \bar{\alpha}_t}}{\sqrt{\bar{\alpha}_t}} 
\epsilon_{\theta}(x_t, t).
\]

\begin{lstlisting}[language=Python]
pre_x_0 = _extract(self.sqrt_recip_alphas_cumprod, t, x_t.shape) * x_t 
          - _extract(self.sqrt_recipm1_alphas_cumprod, t, x_t.shape) 
          * model(x_t, t)
\end{lstlisting}

Substituting this into the posterior mean yields the DDPM update rule:

\[
x_{t-1} =
\frac{1}{\sqrt{\alpha_t}}
\left(
x_t - \frac{1 - \alpha_t}{\sqrt{1 - \bar{\alpha}_t}}
\epsilon_\theta(x_t, t)
\right)
+
\sigma_t z.
\]

Or, in compact form:

\[
x_{t-1} = \mu_{t-1} + \sigma_{t-1} \epsilon.
\]

\subsection{Denoising Equation}

In the implementation, denoising from $x_t$ to $x_{t-1}$ follows:

\[
x_{t-1}
= \mu_{t-1}
+ \text{mask} \cdot \exp\!\left( \tfrac{1}{2} \log \sigma_{t-1}^{2} \right) \cdot \epsilon,
\qquad \epsilon \sim \mathcal{N}(0, I)
\]

where the following components are used:

\paragraph{(1) Log-variance trick}

Instead of directly using $\sigma_{t-1}$, the implementation uses:

\[
\exp\!\left( \tfrac{1}{2} \log \sigma_{t-1}^{2} \right)
\]

which is mathematically identical to $\sigma_{t-1}$.  
This formulation ensures numerical stability during training:

\begin{itemize}
    \item When $\sigma_{t-1}^2$ is extremely small, computing $\sigma_{t-1}$ directly may cause underflow.
	\item Using $\log \sigma_{t-1}^2$ stabilizes gradients and avoids NaNs.
\end{itemize}

\paragraph{(2) Noise mask}

he mask is defined as:

\[
\mathrm{mask} =
\begin{cases}
0, & t = 0, \\
1, & t > 0,
\end{cases}
\]

ensuring that no noise is added at $t=0$.

\begin{lstlisting}[language=Python]
nonzero_mask = ((t != 0).float().view(-1, *([1] * (len(x_t.shape) - 1))))
# no noise when t == 0
\end{lstlisting}

\subsection{Denoising Process}

\textbf{Computing posterior variance}. Given the cumulative noise level $\bar{\alpha}_t$, we obtain
$\bar{\alpha}_{t-1}$ and compute the posterior variance $\sigma^2_{t-1}$ as well as its logarithm $\log \sigma^2_{t-1}$. 

Assuming the noisy input has shape $x_t.\text{shape} = (\text{batch\_size}, \text{channels}, \text{image\_size}, \text{image\_size})$, the following implementation produces a batch of
$\bar{\alpha}_{t-1}$ and $\log \sigma^2_{t-1}$:

\begin{lstlisting}[language=Python]
def q_posterior_mean_variance(self, x_start: torch.FloatTensor,
                              x_t: torch.FloatTensor,
                              t: torch.LongTensor):
    # Compute the mean and variance of the diffusion posterior: q(x_{t-1} | x_t, x_0)
    posterior_mean = (
        self._extract(self.posterior_mean_coef1, t, x_t.shape) * x_start +
        self._extract(self.posterior_mean_coef2, t, x_t.shape) * x_t
    )
    posterior_variance = self._extract(self.posterior_variance, t, x_t.shape)
    posterior_log_variance_clipped = \
        self._extract(self.posterior_log_variance_clipped, t, x_t.shape)
    return posterior_mean, posterior_variance, posterior_log_variance_clipped
\end{lstlisting}

\textbf{Estimating $x_0$}. Using the noisy input $x_t$, the cumulative coefficients $\bar{\alpha}_t$, and the
model-predicted noise $\epsilon_\theta(x_t,t)$, we obtain an estimate of $x_0$.
The result is clipped to the range $\{-1,1\}$.  

This step calls the previous posterior function because computing $x_0$
(x\_start) is required to obtain $\mu_{t-1}$ (model\_mean):

\begin{lstlisting}[language=Python]
def p_mean_variance(self, model, x_t: torch.FloatTensor,
                    t: torch.LongTensor):
    # compute x_0 from x_t and predicted noise: the reverse of q_sample
    pre_x_0 = (
        self._extract(self.sqrt_recip_alphas_cumprod, t, x_t.shape) * x_t -
        self._extract(self.sqrt_recipm1_alphas_cumprod, t, x_t.shape) *
        model(x_t, t)  # pred_noise = model(x_t, t)
    )
    pre_x_0 = torch.clamp(pre_x_0, min=-1., max=1.)  # clip_denoised

    model_mean, posterior_variance, posterior_log_variance = \
        self.q_posterior_mean_variance(pre_x_0, x_t, t)
    return model_mean, posterior_variance, posterior_log_variance
\end{lstlisting}

\textbf{Computing the posterior mean}. Given $x_t$, $x_0$, and the coefficients derived from
$\bar{\alpha}_t$, $\bar{\beta}_t$, and $\bar{\alpha}_{t-1}$, the posterior mean $\mu_{t-1}$ is:

\[
\mu_{t-1}
  = \text{coef}_1 \cdot x_0
  + \text{coef}_2 \cdot x_t .
\]

This is computed in the first function:

\begin{lstlisting}[language=Python]
posterior_mean = (
    self._extract(self.posterior_mean_coef1, t, x_t.shape) * x_start +
    self._extract(self.posterior_mean_coef2, t, x_t.shape) * x_t
)
\end{lstlisting}

\textbf{Sampling $x_{t-1}$.} The transition from $x_t$ to $x_{t-1}$ is computed using the predicted
mean and variance:

\begin{lstlisting}[language=Python]
def p_sample(self, model, x_t: torch.FloatTensor,
             t: torch.LongTensor):
    # denoise_step: sample x_{t-1} from x_t and pred_noise
    model_mean, _, model_log_variance = self.p_mean_variance(model, x_t, t)
    noise = torch.randn_like(x_t)
    nonzero_mask = ((t != 0).float().view(
        -1, *([1] * (len(x_t.shape) - 1)))
    )  # no noise when t == 0
    pred_img = model_mean + nonzero_mask * \
        (0.5 * model_log_variance).exp() * noise
    return pred_img
\end{lstlisting}

\textbf{Full sampling loop.} Starting from Gaussian noise $x_T$, the reverse diffusion process repeats
for \texttt{timesteps} iterations until $x_0$ is obtained:

\begin{lstlisting}[language=Python]
def sample(self, model: nn.Module, image_size,
           batch_size=8, channels=3):
    shape = (batch_size, channels, image_size, image_size)
    device = next(model.parameters()).device

    # start from pure noise
    img = torch.randn(shape, device=device)  # x_T ~ N(0, 1)
    imgs = []

    for i in tqdm(reversed(range(0, self.timesteps)),
                  desc='sampling loop time step',
                  total=self.timesteps):
        t = torch.full((batch_size,), i, device=device,
                       dtype=torch.long)
        img = self.p_sample(model, img, t)
        imgs.append(img.cpu().numpy())
    return imgs
\end{lstlisting}

\section{DDIM}

Overall, DDIM can be viewed as a skip-step sampling version of DDPM. 
It simplifies the sampling equations and greatly accelerates the inference process. 
To simplify the code, we omit the DDPM+DDIM hybrid stochastic term (the optional noise term in the original formulation) 
and retain only the deterministic DDIM formulation.

\subsection{Denoising Equation}

DDPM requires adding a random noise term $\epsilon$ at each step, corresponding to a stochastic SDE sampling process. 
This increases sample diversity but makes sampling slow (typically 100–1000 steps).

\[
x_{t-1} = 
\frac{1}{\sqrt{\alpha_t}}
\left(
    x_t - 
    \frac{1 - \alpha_t}{\sqrt{1 - \bar{\alpha}_t}}
    \cdot \epsilon_\theta(x_t, t)
\right)
+ \sigma_t \cdot \epsilon
=
\sqrt{\bar{\alpha}_{t-1}}\, x_0
+
\sqrt{1 - \bar{\alpha}_{t-1}}
\cdot
\frac{x_t - \sqrt{\bar{\alpha}_t}x_0}{\sqrt{1 - \bar{\alpha}_t}}
+
\sigma_t \cdot \epsilon
\]

DDIM removes the noise term entirely.  
It becomes a deterministic ODE sampler, enabling flexible step scheduling (e.g., 10–50 steps).

\[
x_{t-1} =
\sqrt{\bar{\alpha}_{t-1}}\, x_0
+
\sqrt{1 - \bar{\alpha}_{t-1}}
\cdot
\frac{x_t - \sqrt{\bar{\alpha}_t}x_0}{\sqrt{1 - \bar{\alpha}_t}}
\]

Here, $x_0$ is computed using the model output $\epsilon_\theta(x_t, t)$.

\subsection{Denoising Procedure}

\subsubsection{Sampling Parameters}

\textbf{Starting noise $x_T \sim \mathcal{N}(0, I)$}

\begin{lstlisting}[language=Python]
shape = (batch_size, channels, image_size, image_size)
x_T = torch.randn(shape, device=self.betas.device)
\end{lstlisting}

\textbf{Interval factor $c$ for selecting evenly spaced DDIM steps}

\begin{lstlisting}[language=Python]
c = ddpm_timesteps / ddim_timesteps
\end{lstlisting}

\textbf{Construct DDIM index sequence $T$ and its predecessor sequence $T_{\text{pre}}$}

\begin{lstlisting}[language=Python]
ddim_timestep_seq = torch.tensor(list(range(0, self.timesteps, c))) + 1
ddim_timestep_prev_seq = torch.cat((torch.tensor([0]), ddim_timestep_seq[:-1]))
\end{lstlisting}

\subsubsection{One Denoising Step}

\textbf{Retrieve DDIM timestep indices}

\begin{lstlisting}[language=Python]
t = torch.full((batch_size,), ddim_timestep_seq[i], device=x_T.device, dtype=torch.long)
next_t = torch.full((batch_size,), ddim_timestep_prev_seq[i], device=x_T.device, dtype=torch.long)
\end{lstlisting}

\textbf{Extract $\bar{\alpha}_t$ and $\bar{\alpha}_{t-1}$}

\begin{lstlisting}[language=Python]
alpha_cumprod_t = self._extract(self.alphas_cumprod, t, x_T.shape)
alpha_cumprod_t_prev = self._extract(self.alphas_cumprod, next_t, x_T.shape)
\end{lstlisting}

\textbf{Predict U-Net noise $\epsilon_\theta$}

\begin{lstlisting}[language=Python]
pred_noise = model(x_t, t)
\end{lstlisting}

\textbf{Compute the predicted clean image $x_0$}

\[
x_0 = 
\frac{x_t}{\sqrt{\bar{\alpha}_t}} -
\frac{\sqrt{1 - \bar{\alpha}_t}}{\sqrt{\bar{\alpha}_t}}
\cdot \epsilon_\theta
\]

\begin{lstlisting}[language=Python]
pred_x0 = (xs[-1] - torch.sqrt(1 - alpha_cumprod_t) * pred_noise) / torch.sqrt(alpha_cumprod_t)
pred_x0 = torch.clamp(pred_x0, min=-1., max=1.)
\end{lstlisting}

\textbf{Compute DDIM update to obtain $x_{t-1}$}

\begin{lstlisting}[language=Python]
pred_dir_xt = torch.sqrt(1 - alpha_cumprod_t_prev) * pred_noise
x_t_pre = torch.sqrt(alpha_cumprod_t_prev) * pred_x0 + pred_dir_xt
\end{lstlisting}

\subsection{Full DDIM Sampling Code}

\begin{lstlisting}[language=Python]
# DDIM Inference / Reverse Process
def ddim_sample(self, model, image_size, ddim_timesteps=100, batch_size=8, channels=3):
    shape = (batch_size, channels, image_size, image_size)
    x_T = torch.randn(shape, device=self.betas.device)
    xs = [x_T]

    c = self.timesteps // ddim_timesteps
    ddim_timestep_seq = torch.tensor(list(range(0, self.timesteps, c))) + 1
    ddim_timestep_prev_seq = torch.cat((torch.tensor([0]), ddim_timestep_seq[:-1]))

    for i in tqdm(reversed(range(0, ddim_steps)),
                  desc='ddpm sampling loop time step',
                  total=ddim_steps):

        t = torch.full((batch_size,), ddim_timestep_seq[i], device=x_T.device, dtype=torch.long)
        next_t = torch.full((batch_size,), ddim_timestep_prev_seq[i], device=x_T.device, dtype=torch.long)

        alpha_cumprod_t = self._extract(self.alphas_cumprod, t, x_T.shape)
        alpha_cumprod_t_prev = self._extract(self.alphas_cumprod, next_t, x_T.shape)

        pred_noise = model(xs[-1], t)
        pred_x0 = (xs[-1] - torch.sqrt(1 - alpha_cumprod_t) * pred_noise) / torch.sqrt(alpha_cumprod_t)
        pred_x0 = torch.clamp(pred_x0, min=-1., max=1.)

        pred_dir_xt = torch.sqrt(1 - alpha_cumprod_t_prev) * pred_noise
        x_t_pre = torch.sqrt(alpha_cumprod_t_prev) * pred_x0 + pred_dir_xt

        xs.append(x_t_pre) # Note: We omit variance sampling here.
    return xs
\end{lstlisting}

\section{Model Architecture}

Most open-source diffusion implementations use a complex U-Net with many attention layers and deep ResNet stacks.  
In this work, we intentionally design a \textbf{minimal U-Net}, keeping only the essential components needed for DDPM/DDIM training and sampling.

\subsection{Downsampling Blocks}

We use four downsampling blocks.  
Each block is composed of one or two convolution layers.  
Two types of convolution modules are used:

\begin{itemize}
    \item A convolution layer that preserves spatial resolution.
    \item A convolution layer that downsamples using stride=2.
\end{itemize}

The simplified implementation is shown below:

\begin{lstlisting}[language=Python]
class Upsample(nn.Module):
    def __init__(self, channels, num_groups=32):
        super().__init__()
        self.conv = nn.Conv2d(channels, channels, kernel_size=3, padding=1)
        self.num_groups = num_groups

    def forward(self, x):
        x = F.interpolate(x, scale_factor=2, mode="nearest")
        x = self.conv(x)
        return x
        
down_block1 = nn.Conv2d(io_channels, model_channels, kernel_size=3, padding=1)
down_block2 = Downsample(model_channels)
\end{lstlisting}

\subsection{Middle Block}

The middle block contains one residual block composed of two convolution layers:

\begin{lstlisting}[language=Python]
class ResidualBlock(nn.Module):
    def __init__(self, in_channels, out_channels):
        super().__init__()
        self.conv1 = nn.Conv2d(in_channels, out_channels, kernel_size=3, padding=1)
        self.conv2 = nn.Conv2d(out_channels, out_channels, kernel_size=3, padding=1)
        self.shortcut = (nn.Conv2d(in_channels, out_channels, kernel_size=1)
                         if in_channels != out_channels else nn.Identity())

    def forward(self, x):
        h = F.relu(F.group_norm(self.conv1(x), num_groups=32))
        h = F.relu(F.group_norm(self.conv2(h), num_groups=32))
        return h + self.shortcut(x)
\end{lstlisting}

We also inject time information $t$ into the middle block.  
This follows the standard positional encoding approach used in diffusion models.

\subsection{Time Embedding Module}

\subsubsection{Code}

The time embedding is implemented using sinusoidal functions:

\begin{lstlisting}[language=Python]
def timestep_embedding(t, dim, max_period=10000):
    freqs = torch.exp(-math.log(max_period) *
                      torch.arange(start=0, end=dim // 2, dtype=torch.float32)
                      / (dim // 2)).to(device=t.device)
    args = t[:, None].float() * freqs[None]
    embedding = torch.cat([torch.cos(args), torch.sin(args)], dim=-1)
    return embedding
\end{lstlisting}

Input:  
\begin{itemize}
    \item t.shape = batchsize
    \item dim.shape = dimension of embedding (layer channels)
\end{itemize}

Output:  
\[
\text{embedding} \in \mathbb{R}^{B \times \text{dim}}
\]

\subsubsection{Computation Example}

Assume:

- $\text{timesteps} = [0,1,2,3]$
- $\text{dim} = 8$

\paragraph{1. Compute half dimension}
\[
\text{half} = \frac{8}{2} = 4
\]

\paragraph{2. Compute frequency values}
\[
\text{freqs}[i] = e^{ -\log(10000) \cdot \frac{i}{4} }
\]

\[
\text{freqs} = [1.0,\; 0.1,\; 0.01,\; 0.001]
\]

\paragraph{3. Compute args}
\[
\text{args}[i,j] = t_i \cdot \text{freqs}_j
\]

\[
\text{args} =
\begin{bmatrix}
0.00 & 0.00 & 0.00 & 0.000 \\
1.00 & 0.10 & 0.01 & 0.001 \\
2.00 & 0.20 & 0.02 & 0.002 \\
3.00 & 0.30 & 0.03 & 0.003
\end{bmatrix}
\]

\paragraph{4. Cosine part}
\[
\cos(\text{args}) = 
\begin{bmatrix}
1.0000 & 1.0000 & 1.0000 & 1.0000 \\
0.5403 & 0.9950 & 0.9999 & 1.0000 \\
-0.4161 & 0.9801 & 0.9998 & 1.0000 \\
-0.9899 & 0.9553 & 0.9996 & 1.0000
\end{bmatrix}
\]

\paragraph{5. Sine part}
\[
\sin(\text{args}) = 
\begin{bmatrix}
0.0000 & 0.0000 & 0.0000 & 0.0000 \\
0.8415 & 0.0998 & 0.0100 & 0.0010 \\
0.9093 & 0.1987 & 0.0200 & 0.0020 \\
0.1411 & 0.2955 & 0.0300 & 0.0030
\end{bmatrix}
\]

\paragraph{6. Concatenated embedding}

\[
\text{embedding} =
\begin{bmatrix}
1.0000 & 1.0000 & 1.0000 & 1.0000 & 0.0000 & 0.0000 & 0.0000 & 0.0000 \\
0.5403 & 0.9950 & 0.9999 & 1.0000 & 0.8415 & 0.0998 & 0.0100 & 0.0010 \\
-0.4161 & 0.9801 & 0.9998 & 1.0000 & 0.9093 & 0.1987 & 0.0200 & 0.0020 \\
-0.9899 & 0.9553 & 0.9996 & 1.0000 & 0.1411 & 0.2955 & 0.0300 & 0.0030
\end{bmatrix}
\]

\paragraph{Summary}

The time embedding represents $t$ using multi-scale sinusoidal patterns:

\[
[\cos(t \cdot \alpha_i),\ \sin(t \cdot \alpha_i)] \quad \alpha_i \in \{1,\; 0.1,\; 0.01,\; ...\}
\]

\subsubsection{Time Embedding Injection}

The embedding is added to the middle layer via broadcasting:

\begin{lstlisting}[language=Python]
noise_embedding = nn.Linear(model_channels, model_channels*2)
middle = middle_block(x)
noise_t = F.relu(self.noise_embedding(timestep_embedding(t,self.model_channels)))
middle = middle + noise_t[:, :, None, None]
\end{lstlisting}

\subsection{Complete U-Net Summary}

Key features:

\begin{itemize}
    \item Downsampling blocks concatenate symmetric upsampling blocks.
    \item Only one middle block is used.
    \item Time embedding is injected only at the middle layer.
    \item No attention layers are used.
\end{itemize}

The full implementation is shown below:

\begin{lstlisting}[language=Python]
class UNetModel(nn.Module):
    def __init__(self, io_channels=3, model_channels=128):
        super().__init__()
        self.model_channels = model_channels

        self.down_block1 = nn.Conv2d(io_channels, model_channels, 3, padding=1)
        self.down_block2 = Downsample(model_channels)
        self.down_block3 = nn.Conv2d(model_channels, model_channels*2, 3, padding=1)
        self.down_block4 = Downsample(model_channels*2)

        self.middle_block = ResidualBlock(model_channels*2, model_channels*2)
        self.noise_embedding = nn.Linear(model_channels, model_channels*2)

        self.up_block1 = Upsample(model_channels*2)
        self.up_block2 = nn.Conv2d(model_channels*2, model_channels, 3, padding=1)
        self.up_block3 = Upsample(model_channels)
        self.up_block4 = nn.Conv2d(model_channels, io_channels, 3, padding=1)

    def forward(self, x, t):
        x1 = F.relu(F.group_norm(self.down_block1(x), 32))
        x2 = F.relu(F.group_norm(self.down_block2(x1), 32))
        x3 = F.relu(F.group_norm(self.down_block3(x2), 32))
        x4 = F.relu(F.group_norm(self.down_block4(x3), 32))

        middle = self.middle_block(x4)
        noise_t = F.relu(self.noise_embedding(timestep_embedding(t,self.model_channels)))
        middle = middle + noise_t[:, :, None, None]

        x5 = F.relu(F.group_norm(self.up_block1(middle + x4), 32))
        x6 = F.relu(F.group_norm(self.up_block2(x5 + x3), 32))
        x7 = F.relu(F.group_norm(self.up_block3(x6 + x2), 32))
        out = self.up_block4(x7 + x1)
        return out
\end{lstlisting}

\section{Classifier-Free Guidance (Conditional Generation)}

Conditional generation has been extensively explored in Generative Adversarial Networks (GANs), with approaches such as CGAN \cite{cgan}, ACGAN \cite{acgan}, and InfoGAN \cite{infogan}. These methods generate samples conditioned on semantic labels, and the labels are provided as additional inputs during training.

In general, conditional generation methods can be categorized into three types:

\begin{itemize}
    \item \textbf{Supervision:}  
    The model predicts the class label in addition to generating the image, and multi-label classification losses are added during training.

    \item \textbf{Weak supervision:}  
    Labels are only used as additional features and embedded into image features (e.g., added element-wise), but the training objective remains unchanged. No extra output heads or additional loss terms are introduced.

    \item \textbf{Unsupervised (pseudo-label based):}  
    Pseudo labels are automatically generated according to some underlying structure (e.g., clustering assignments or contrastive-learning-based representations), and the training requires corresponding auxiliary losses such as classification, clustering, or contrastive losses.
\end{itemize}

Classifier-Free Guidance belongs to the weakly supervised category. The label is injected into the model as an additional feature, similar to the timestep embedding. Specifically, the label embedding is added to the U-Net middle block and broadcast across spatial dimensions. The conditioning can be written as:

\[
x = x + t_{\mathrm{emb}} + c_{\mathrm{emb}},
\]

where $t_{\mathrm{emb}}$ is the timestep embedding and $c_{\mathrm{emb}}$ is the embedding vector corresponding to the class label.

Since labels are discrete, an \texttt{Embedding} layer is used instead of a fully connected layer. The corresponding PyTorch implementation is shown below:

\begin{lstlisting}[language=Python]
class UNetModel(nn.Module):
        ...
        self.down_block4  = Downsample(model_channels*2)

        self.middle_block = ResidualBlock(model_channels*2, model_channels*2)  # middle block
        self.noise_embedding = nn.Linear(model_channels, model_channels*2)     # noise block
        self.class_emb = nn.Embedding(class_num, model_channels*2)

        self.up_block1 = Upsample(model_channels*2)  # up blocks
        ...
        
    def forward(self, x, t, label=None):
        ...
        x4 = F.relu(F.group_norm(self.down_block4(x3), num_groups=32))

        middle = self.middle_block(x4)  # Middle block
        noise_t = F.relu(self.noise_embedding(timestep_embedding(t, self.model_channels)))
        c_emb = F.relu(self.class_emb(label))
        middle = middle + noise_t[:, :, None, None] + c_emb[:, :, None, None]

        x5 = F.relu(F.group_norm(self.up_block1(middle + x4), num_groups=32))  # Decode-Upsampling
        ...
        out = self.up_block4(x7 + x1)

        return out
\end{lstlisting}

Here, \textbf{c\_emb} denotes the embedding vector corresponding to the class label.

\section{Experiment Results}

We evaluate our model on MNIST, Fashion-MNIST, and CIFAR-10.

\subsection{Hyperparameters}

\vspace{5pt}
\textbf{MNIST / Fashion-MNIST}:

\begin{itemize}
    \item Epochs: 200
    \item Timesteps: 300
\end{itemize}

\textbf{CIFAR-10}:

\begin{itemize}
    \item Epochs: 500
    \item Timesteps: 1000
    \item Deeper model: additional CNN layers in Down, Middle, and Up blocks
\end{itemize}

\subsection{DDPM Results}

\subsubsection{MNIST}
\begin{figure}[htbp]
    \centering
    \begin{subfigure}{0.47\linewidth}
        \includegraphics[width=\linewidth]{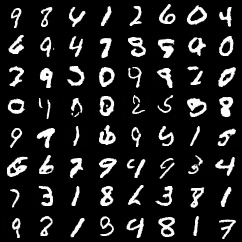}
        \caption{Random samples}
    \end{subfigure}
    \hfill
    \begin{subfigure}{0.47\linewidth}
        \includegraphics[width=\linewidth]{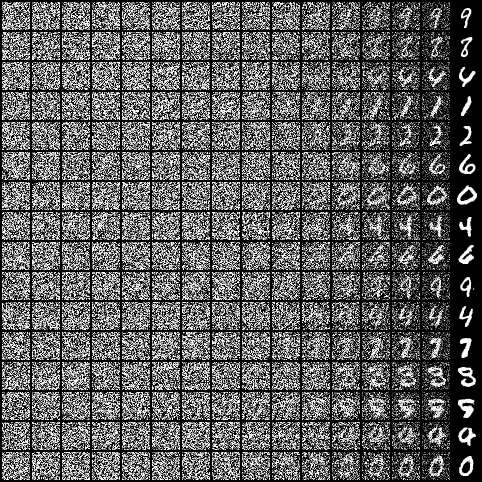}
        \caption{Denoising random samples}
    \end{subfigure}
    \caption{DDPM results on MNIST.}
    \label{ddpm-mnist}
\end{figure}

MNIST results on 200 epochs are show Fig. \ref{ddpm-mnist}.

\subsubsection{Fashion-MNIST}
\begin{figure}[H]
    \centering
    \begin{subfigure}{0.46\linewidth}
        \includegraphics[width=\linewidth]{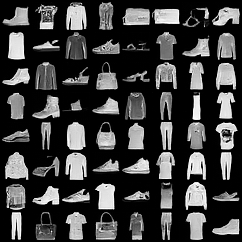}
        \caption{Random samples}
    \end{subfigure}
    \hfill
    \begin{subfigure}{0.48\linewidth}
        \includegraphics[width=\linewidth]{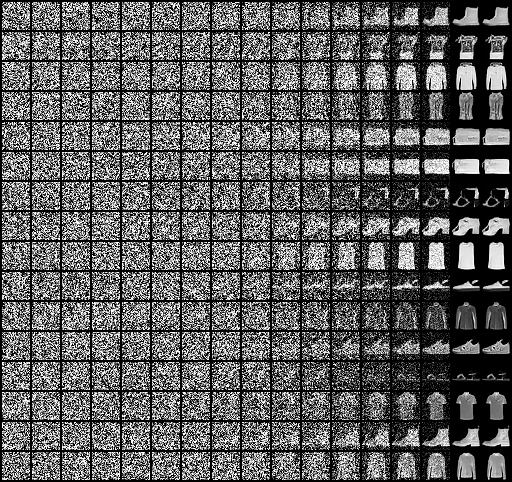}
        \caption{Denoising random samples}
    \end{subfigure}
    \caption{DDPM results on Fashion-MNIST.}
    \label{ddpm-fashion-mnist}
\end{figure}

Fashion-MNIST results on 200 epochs are show in Fig. \ref{ddpm-fashion-mnist}.

\subsubsection{CIFAR-10}

\begin{figure}[htbp]
    \centering
    \begin{subfigure}{0.49\linewidth}
        \includegraphics[width=\linewidth]{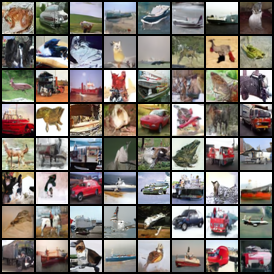}
        \caption{Random samples on 490 epochs}
    \end{subfigure}
    \hfill
    \begin{subfigure}{0.49\linewidth}
        \includegraphics[width=\linewidth]{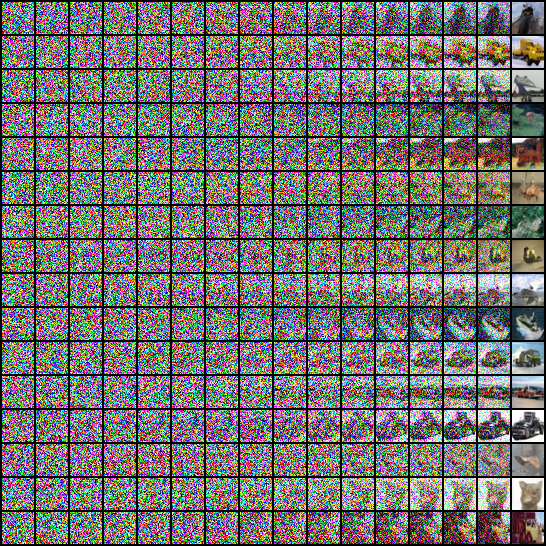}
        \caption{Denoising random samples on 470 epochs}
    \end{subfigure}
    \caption{DDPM results on CIFAR-10.}
    \label{ddpm-cifar-10}
\end{figure}

CIFAR-10 is significantly more challenging due to large inter-class variation and low resolution (32×32).  
Despite using a relatively small model, we increase network depth for CIFAR-10 training (see our release code for details).  Results at 470-490 epochs are shown in Fig. \ref{ddpm-cifar-10}.

\subsection{DDIM Results}

We evaluate DDIM with only 50 sampling steps.  
Empirically, as few as 10 steps already produce reasonable images, although still worse performance than DDPM.

Fashion-MNIST results are show in Fig. \ref{ddim-fashion-mnist}:

\begin{figure}[H]
    \centering
    \begin{subfigure}{0.45\linewidth}
        \includegraphics[width=\linewidth]{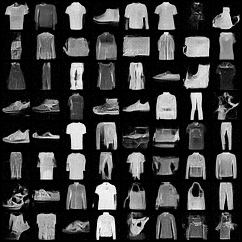}
        \caption{Random samples}
    \end{subfigure}
    \hfill
    \begin{subfigure}{0.5\linewidth}
        \includegraphics[width=\linewidth]{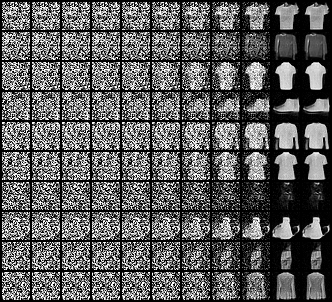}
        \caption{Denoising random samples}
    \end{subfigure}
    \caption{DDIM (50 steps) results on Fashion-MNIST.}
    \label{ddim-fashion-mnist}
\end{figure}

\subsection{Classifier-Free Guidance}

With Classifier-Free Guidance added to the model, we perform label-conditioned generation  
using DDIM with only 10 sampling timesteps.

\begin{figure}[H]
    \centering
    \begin{subfigure}{0.45\linewidth}
        \includegraphics[width=\linewidth]{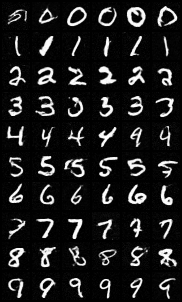}
        \caption{MNIST conditional random samples}
    \end{subfigure}
    \hfill
    \begin{subfigure}{0.45\linewidth}
        \includegraphics[width=\linewidth]{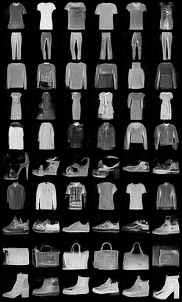}
        \caption{Fashion-MNIST conditional random samples}
    \end{subfigure}
    \caption{Classifier-Free Guided (CGF) conditional generation (DDIM-10 timesteps).}
\end{figure}

\section{Note on Original Version}

This report was originally written in Chinese on 2025-04-02.  The original version is available on the following platforms:  
\begin{itemize}
  \item WeChat public account: \url{https://mp.weixin.qq.com/s/C0zedCz99P48IxgpL6XDyQ}
  \item Zhihu Q\&A: \url{https://www.zhihu.com/question/567829973/answer/1890784793505032031}
\end{itemize}

This report constitutes an English translation and reorganization of the original work, with minor additions, aiming to broaden its accessibility and support researchers who may find it valuable. 

\bibliography{myReference}
\end{document}